\documentclass[journal]{IEEEtran}
\usepackage{times}
\usepackage{epsfig}
\usepackage{graphicx}
\usepackage{amsmath}
\usepackage{amssymb}
\usepackage{cite}

\usepackage{hyperref}

\usepackage{booktabs}
\usepackage{gensymb}
\usepackage{multirow}
\usepackage[norule,symbol,perpage]{footmisc}

\def\pt{p_\textrm{t}}
\def\at{\alpha_\textrm{t}}

\def\CE{\textrm{CE}}
\def\FL{\textrm{FL}}

\newcommand{\eqnnm}[2]{\begin{equation}\label{eq:#1}#2\end{equation}\ignorespaces}

\begin{document}

\title{Can Your Face Detector Do Anti-spoofing? \\ Face Presentation Attack Detection with a Multi-Channel Face Detector}

\author{Anjith George and S\'ebastien Marcel \\
Idiap Research Institute \\
Rue Marconi 19, CH - 1920, Martigny, Switzerland \\
{\tt\small  \{anjith.george, sebastien.marcel\}@idiap.ch  }
}

\maketitle

\begin{abstract}
In a typical face recognition pipeline, the task of the face detector is to localize the face region. However, the face detector localizes regions that look like a face, irrespective of the liveliness of the face, which makes the entire system susceptible to presentation attacks. In this work, we try to reformulate the task of the face detector to detect \textit{real} faces, thus eliminating the threat of presentation attacks. While this task could be challenging with visible spectrum images alone, we leverage the multi-channel information available from off the shelf devices (such as color, depth, and infrared channels) to design a multi-channel face detector. The proposed system can be used as a \textit{live-face detector} obviating the need for a separate presentation attack detection module, making the system reliable in practice without any additional computational overhead. The main idea is to leverage a single-stage object detection framework, with a joint representation obtained from different channels for the PAD task. We have evaluated our approach in the multi-channel \textit{WMCA} dataset containing a wide variety of attacks to show the effectiveness of the proposed framework.
\end{abstract}

\section{Introduction}

Face recognition technology has become ubiquitous these days, thanks to the advance in deep learning based methods \cite{jain2011handbook}. However, the reliability of face recognition systems (FRS) in the presence of attacks is poor in practical situations, mainly due to the vulnerability against presentation attacks (a.k.a spoofing attacks) \cite{handbook2}, \cite{ISO}. Presenting artifacts like a photograph or video in front of the camera could be enough to fool unprotected FR systems. The artifact used for this attack is known as a presentation attack instrument (PAI).

Presentation attack detection (PAD) tries to protect the FR systems against such attacks. Majority of the PAD methods proposed in the literature focuses on the detection of 2D attacks. Most of these methods use either feature-based methods or CNN based approaches using visible spectrum images. However, they do not emulate a realistic scenario of encountering realistic 3D attacks. Recently, there has been a lot of works focusing on the detection of a wide variety of attacks, including 2D, 3D, and partial attacks. Multi-channel methods have been proposed as a possible alternative to deal with real-world scenarios with a wide variety of attacks \cite{george_mccnn_tifs2019, nikisins2019domain}. Usage of multiple channels makes it harder to fool the PAD systems.

\begin{figure}[t]
     \centering
         \includegraphics[width=0.99\linewidth]{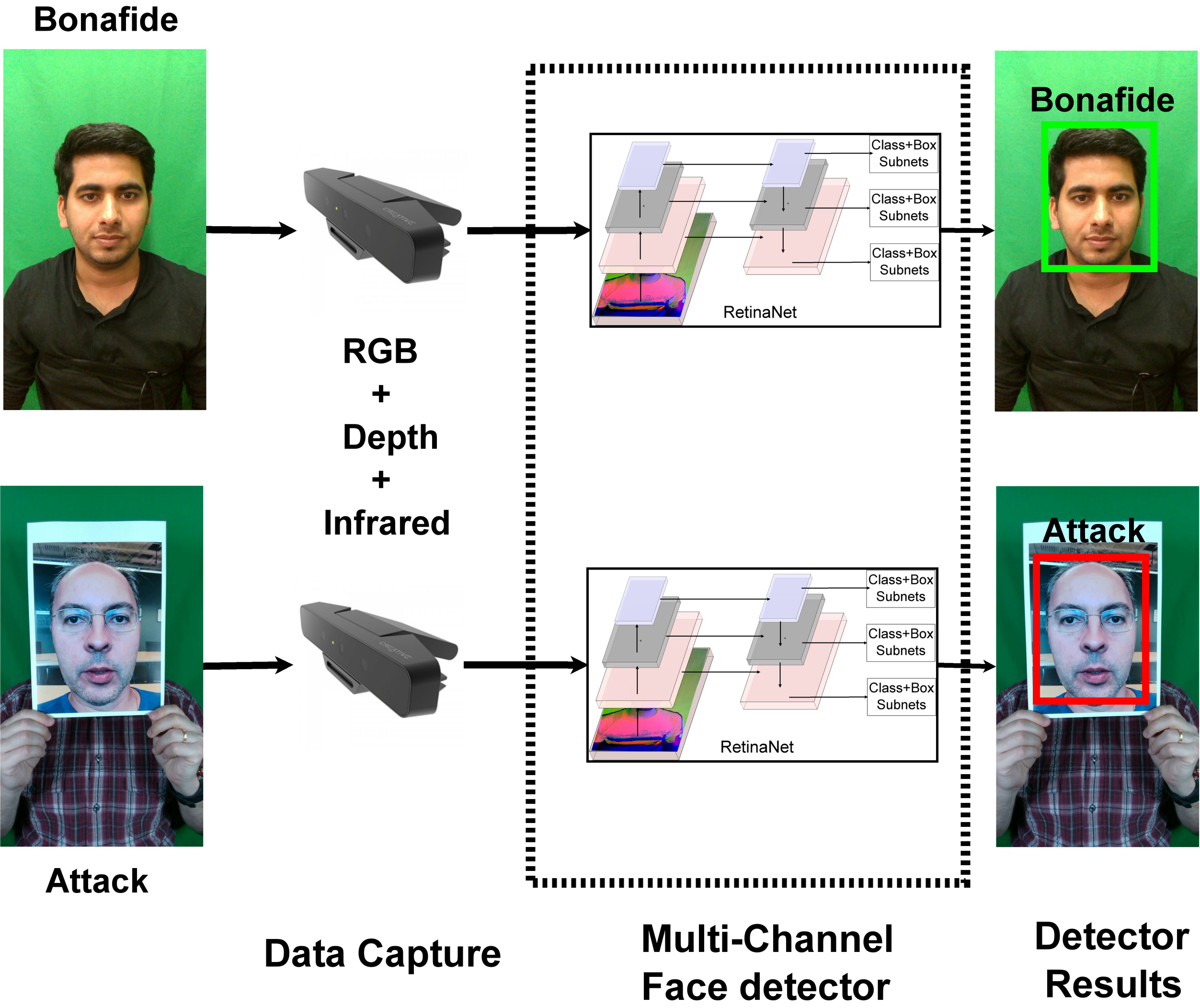}

\caption{Schematic of the proposed framework. The camera captures images in color, depth and infrared channels (all channel coming from a single commercially available device) and the RetinaNet based multi channel face detector performs simultaneous face detection and presentation attack detection using the composite image from these channels. The framework can be adapted to work with RGB-D combination just by stacking RGB and Depth channels at the input side (to make it work with OpenCV AI Kit (OAK-D) \cite{oak_d} and Kinect \cite{zhang2012microsoft}).}
\label{fig:framework}
\end{figure}

For reliable usage of face recognition systems, they must be equipped with a presentation attack detection module. The PAD module can act before, after, or together with the face recognition module.

Typically face recognition frameworks consists of a preprocessing stage including face detection and alignment, followed by the actual face recognition task. In the proposed PAD framework, we propose to use a multi-channel face detector \textit{as} the PAD system. In this way, the preprocessing stage for the face recognition system itself can perform PAD, thanks to the multi-channel information. Furthermore, the face detectors used typically are also CNN based, so if we can replace this face detector with the proposed module, it should not increase the overall complexity while adding PAD capabilities, simplifying the overall face recognition pipeline by removing the redundancy.

Presentation attack detection is achieved by changing the task of the face detector from \textit{detecting faces} to \textit{detecting bonafide faces}. This is a hard challenge when only RGB channels are used and can result in a lot of detection errors. However, the proposed face detection based PAD framework (Fig. \ref{fig:framework}) leverage the multi-channel information to discriminate between \textit{bonafide} and presentation attacks.

Another advantage of the proposed face detection-based framework is the efficient use of the background. Most of the common PAD frameworks utilize only the facial region and completely ignores the information from the background. Notably, in a multi-channel framework, background could give beneficial negative samples to make the system robust. The proposed framework can be configured based on the availability of data in different ways. It can be trained as a purely one class model when no negative samples are available, i.e, when only \textit{bonafide} samples are available. The \textit{bonafide} face location can be used as positive samples, and the entire background can be used as negative samples. When attacks are available, the face detector can be trained to classify between \textit{bonafide} and the attack classes. It is also possible to train the face detector as a multi-class classifier when we have the labels for different kinds of attack classes. While training the face detector, a lot of negative samples contribute to the loss function. To avoid this, we use the focal loss formulation, which focuses on the hard examples. If the face detection network is trained only on \textit{bonafide} samples, most of the negative samples (patches) would be easy to classify. The attack class images can be used as hard negative samples. This introduces greater flexibility as the face detector can be trained with different configurations, for instance, as a one-class classifier, either with only \textit{bonafide} class, or with the addition of attack classes too in training providing the CNN based face detector with harder examples, or as multi class classifier. This could be useful in the cases where only a limited amount of attacks are available in training.

To the best knowledge of the authors, this is the first work using a multi-channel face detector for the task of face presentation attack detection. The main contributions of this work are listed below.

\begin{itemize}
\item Proposes a novel framework for PAD, using a multi-channel face detector, performing simultaneous face localization and presentation attack detection.
\item The channels used in this work comes from an affordable consumer-grade camera, which makes it easy to deploy this system in real-world conditions.
\item The proposed algorithm can be used as a preprocessing stage in face recognition systems, thus reducing the redundancy of an additional PAD system in the face recognition pipeline.
\end{itemize}

\section{Related work}
Most of the prevailing literature in face presentation attack detection deals with the detection of 2D attacks using visible spectrum images.  Majority of them depend on the quality degradation of the recaptured samples for PAD. Methods such as motion patterns \cite{anjos2011counter}, Local Binary Patterns (LBP) \cite{boulkenafet2015face}, image quality features \cite{galbally2014image}, and image distortion analysis \cite{wen2015face} are examples of feature based PAD methods. Also, there are several CNN based methods achieving state of the art performance \cite{liu2018learning,george-icb-2019, atoum2017face}. While there has been a lot of work in the detection of 2D presentation attacks, the assumptions of quality degradation during recapture does not hold for attacks such as realistic silicone masks and partial attacks. Here we limit the discussion to recent and representative methods that handle a wide variety of 2D and 3D attacks.

With the ever-improving quality of attacks, visible spectrum images alone may not suffice for detecting presentation attacks. This problem becomes more severe when there is a wide variety of possible 2D and 3D presentation attacks possible. Multi-channel and multi-spectral methods have been proposed as a solution for this problem \cite{raghavendra2017extended}, \cite{steiner2016reliable}.

Raghavendra \textit{et al}. \cite{raghavendra2017extended} presented an approach to use complementary information from different channels using a multi-spectral PAD framework. Their method used a fusion of wavelet-based features. Score level fusion achieved better performance as compared to feature fusion in detecting attacks prepared using different kinds of printers. Erdogmus and Marcel \cite{erdogmus2014spoofing} showed that 3D masks could fool the PAD systems easily. By combining the LBP features from color and depth channels, they could achieve good performance in the 3DMAD dataset.

Steiner \textit{et al}. \cite{steiner2016reliable} introduced multi-spectral SWIR image-based PAD method, capturing four different wavelengths - 935\textit{nm}, 1060\textit{nm}, 1300\textit{nm} and 1550\textit{nm}. Their method essentially consisted of a skin level classifier in a predefined Region Of Interest (ROI) where the skin was expected to be present. They trained a pixel-level SVM to classify each pixel as skin or not. The percentage of skin detections in the ROI was used as the PAD score. Their method achieved 99.28\% accuracy in pixel-level skin classification.

In \cite{dhamecha2013disguise}, the authors combined visible and thermal image patches for PAD. The patches used for the subsequent face recognition stage were selected by first classifying the patches as either \textit{bonafide} or attacks.

Bhattacharjee \textit{et al}. \cite{Bhattacharjee:256262} showed the vulnerability of CNN based face recognition systems against 3D masks. They also proposed
simple thermal-based features for PAD. Further, in \cite{bhattacharjee2017you}, they presented preliminary experiments on the use of additional channels for PAD.

In \cite{george_mccnn_tifs2019}, George \textit{et al}. presented a multi-channel face presentation attack detection framework. Their approach consisted of extending a pretrained face recognition network to accept multiple channels and to adapt a minimal number of layers to prevent overfitting. The proposed method achieved an error rate of 0.3 \% in the challenging \textit{WMCA} dataset using color, depth, infrared and thermal channels. Further several other works have reported improved performance with the use of multi-channel methods \cite{heusch2020deep,george-tifs-2020}.

Many multi-channel datasets have been made available for PAD recently. However, the variety of attacks is rather limited in most of the available databases.

Typical PAD frameworks perform a preprocessing stage involving face detection with RGB channel, followed by alignment and binary classification for PAD. The PAD classifier, for most of them, has the form of a binary classifier acting on the detected face regions which might result in poor performance in unseen attack scenarios. Such frameworks ignores the useful information coming from the background regions completely. Moreover, the PAD module adds additional complexity to the face recognition pipeline.

\section{Proposed approach}

In the proposed approach, we use an object detection framework for localizing face and classifying between \textit{bonafide} and \textit{non-face} (attacks) labels. Specifically, we formulate the problem of multi-channel presentation attack detection as a two-class object detection problem. We leverage a single-stage object detection framework for this purpose. The task for the proposed object detector is to detect the presence of `\textit{bonafide}' or `\textit{non-face}' classes. While training, face locations detected by the RGB face detector \cite{zhang2016joint} is used as the ground truth bounding box locations. The class labels of the bounding boxes is considered as `\textit{bonafide}' class for real samples and all other attack classes are grouped as `\textit{non-face}' class. In the evaluation phase, the confidence of the \textit{bonafide} class is used for the scoring. If no object classes (`\textit{bonafide}' or `\textit{non-face}') are detected, then the sample is considered as an attack which was not seen in training. We use the RetinaNet \cite{lin2017focal} as the base architecture for the proposed multi-channel face detector, which obtained state of the art performance in object detection tasks.

\subsection{Preprocessing}
The data used in the network comprises of color, depth, and infrared channels. The color channel is converted to gray-scale. The raw depth and infrared channels available from the hardware are in 16-bit format. We first normalize these channels with Median Absolute Deviation (MAD) based normalization to convert them to 8 bit \cite{nikisins2019domain}.

This stage is detailed as follows, let $\mathbf{I}$ be the image; first,  a vector $\mathbf{v}$ containing non-zero elements of $\mathbf{I}$ is obtained to reduce the effect of background holes and dead pixels. The MAD value is computed as follows,
\begin{equation}
\mathbb{MAD} = median(|\mathbf{v}-median(\mathbf{v})|)
\end{equation}
Once the $\mathbb{MAD}$ is computed, the image can be normalized to 8 bit as:
\begin{equation}
\hat{\mathbf{I}}_{i,j} = \frac{(\mathbf{I}_{i,j} - median(\mathbf{v}) + \sigma \cdot \mathbb{MAD})}{2\cdot\sigma\cdot\mathbb{MAD}} \cdot (2^8-1),
\end{equation}

Values $i$ and $j$ denote the coordinates of the pixels in the image. The $\sigma$ value used in our experiments was four.

\begin{figure}[t]
     \centering
         \includegraphics[width=0.99\linewidth]{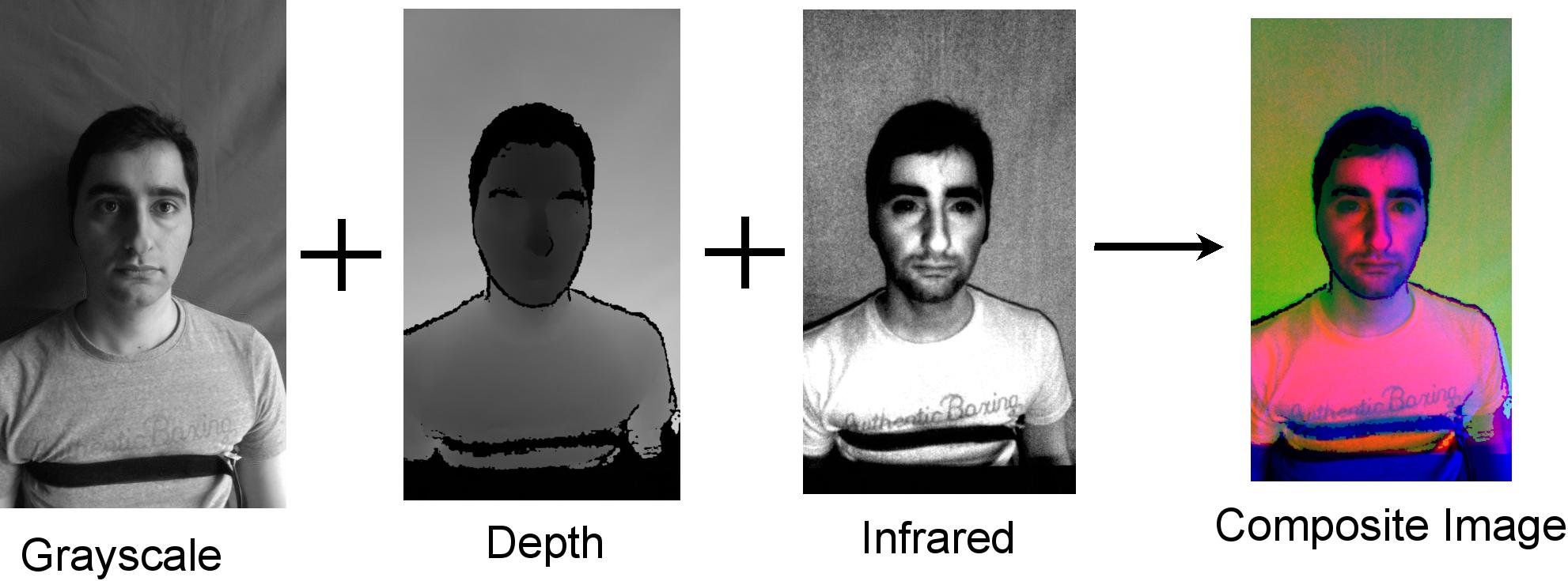}

\caption{The composite image is formed from gray-scale, depth and infrared channels by stacking the normalized images.}
\label{fig:composite}
\end{figure}

Once all the channels are available in 8-bit format, they are concatenated to form the grayscale-depth-infrared composite image, which is used as the input in the subsequent CNN pipeline.
An example of the composite \textit{bonafide} image is shown in Fig. \ref{fig:composite}

\subsection{Network architecture and loss function}

The objective is to treat PAD problem as an object detection problem. In literature, there are several two-stage detectors that can do object detection by first generating the candidate locations and then classifying the candidate locations to get object labels. However, one stage object detectors are advantageous due to their simple design and faster inference. The RetinaNet \cite{lin2017focal} architecture is a one-stage object detection architecture, which is simpler and quicker at the inference stage. One of the main issues with one stage detectors is the poor performance due to heavy class imbalance. Out of a large number of candidate locations, only a few of them contain objects. This could result in poor models as the loss includes a lot of contributions from the background. In RetinaNet \cite{lin2017focal}, this issue is handled by introducing a new loss function called focal loss. This is done by down-weighting easy examples so that their contribution is less in the loss function. In summary, focal loss naturally favors the training on hard instances.

\subsubsection{Architecture}

We use the standard RetinaNet architecture built from a backbone network as the object detector. In our implementation, we used  Feature Pyramid Network \cite{lin2017feature} with ResNet-18 \cite{he2016deep} backbone. There are two subnetworks; one is tasked with regressing the bounding box and the other one is performing the classification. In our case, the classifier needs to classify the presence of `\textit{bonafide}' or `\textit{non-face}' categories. The task of the regression network is to predict the `\textit{bonafide}' (or `\textit{non-face}' ) bounding box. Instead of RGB images, we use the composite image created in the preprocessing stage as the input to the network.

\subsubsection{Loss function}

As shown earlier, the RetinaNet architecture consists of two subnetworks, one for landmark regression and one for classification. For the object classification, the typically used cross entropy loss (CE) has the following form:

\begin{equation}
{\CE(p,y) = \begin{cases} -\log(p) &\text{if $y = 1$}\\
 -\log (1 - p) &\text{otherwise.}\end{cases}}
\end{equation}

In the above equation, $y \in \{\pm1\}$ specifies the label, $p \in [0,1]$ is probability computed by the network for the class with label $y=1$. We define $\pt$ as:

 \eqnnm{pt}{\pt=\begin{cases} p &\text{if $y = 1$}\\ 1 - p &\text{otherwise,}\end{cases}}
so that we can rewrite the loss as  $\CE(p,y) = \CE(\pt) = - \log (\pt)$.

After adding the modulation factor {\small $(1 - \pt)^\gamma$}, and  $\alpha$-balancing the expression for focal loss (FL) becomes \cite{lin2017focal}:

 \eqnnm{flalpha}{\FL(\pt) = - \at (1 - \pt)^\gamma \log (\pt).}

The classification head of the network is trained with this loss function which reduces the effect of easily classifiable background patches in the loss function.

\subsubsection{Scoring method}

The network described above is trained as an object detector to localize and classify face. Here the notation `\textit{face}' denotes the location of face irrespective of the class label. The possible classes are `\textit{bonafide}' and `\textit{non-face}'.
The `\textit{non-face}' category might contain a wide variety of attacks, and its possible that in some cases, no objects are detected. However, for the PAD system, we need to get a score, which is defined as the probability of a \textit{bonafide} class. The network outputs a classification label and class probability for the detected class, above a detection threshold. Once this information is available, the scoring is performed as follows. If the object class predicted is `\textit{bonafide}' then PAD score is computed as $score=P_{bonafide}$ where $P_{bonafide}$ is the probability of `\textit{bonafide}' class. If the object class returned is `\textit{non-face}' then $score=1-P_{non-face}$ where $P_{non-face}$ is the probability of the `\textit{non-face}' class. In the case where no objects are detected, as the case when an unseen attack is present, the sample is considered as an attack and given a predefined low score (i.e., considered as an attack).

\subsubsection{Implementation details}

We used a standard RetinaNet architecture with ResNet-18 \cite{he2016deep} backbone for our experiments. Pretrained weights of a model trained on ImageNet \cite{deng2009imagenet} was used to initialize the ResNet blocks. To train the face detector, $train$ set of the PAD database was used for training, and the $dev$ set was used for validation.  The model was trained as a standard object detector with two classes (`\textit{bonafide}' and `\textit{non-face}' classes).
The face bounding box used in the training was obtained from MTCNN algorithm \cite{zhang2016joint} on color channel images. Adam Optimizer \cite{kingma2014adam} with a learning rate of $2\times10^{-5}$ was used in the training. The model was trained for 50 epochs with ten frames from each video on a GPU grid. All the layers were adapted during the network training. The model corresponding to minimum validation loss was selected for the evaluation. The implementation was done with PyTorch \cite{paszke2017automatic}.

\section{Experiments and Results}

\subsection{Databases used}

\begin{figure}[t]
     \centering
         \includegraphics[width=1\linewidth]{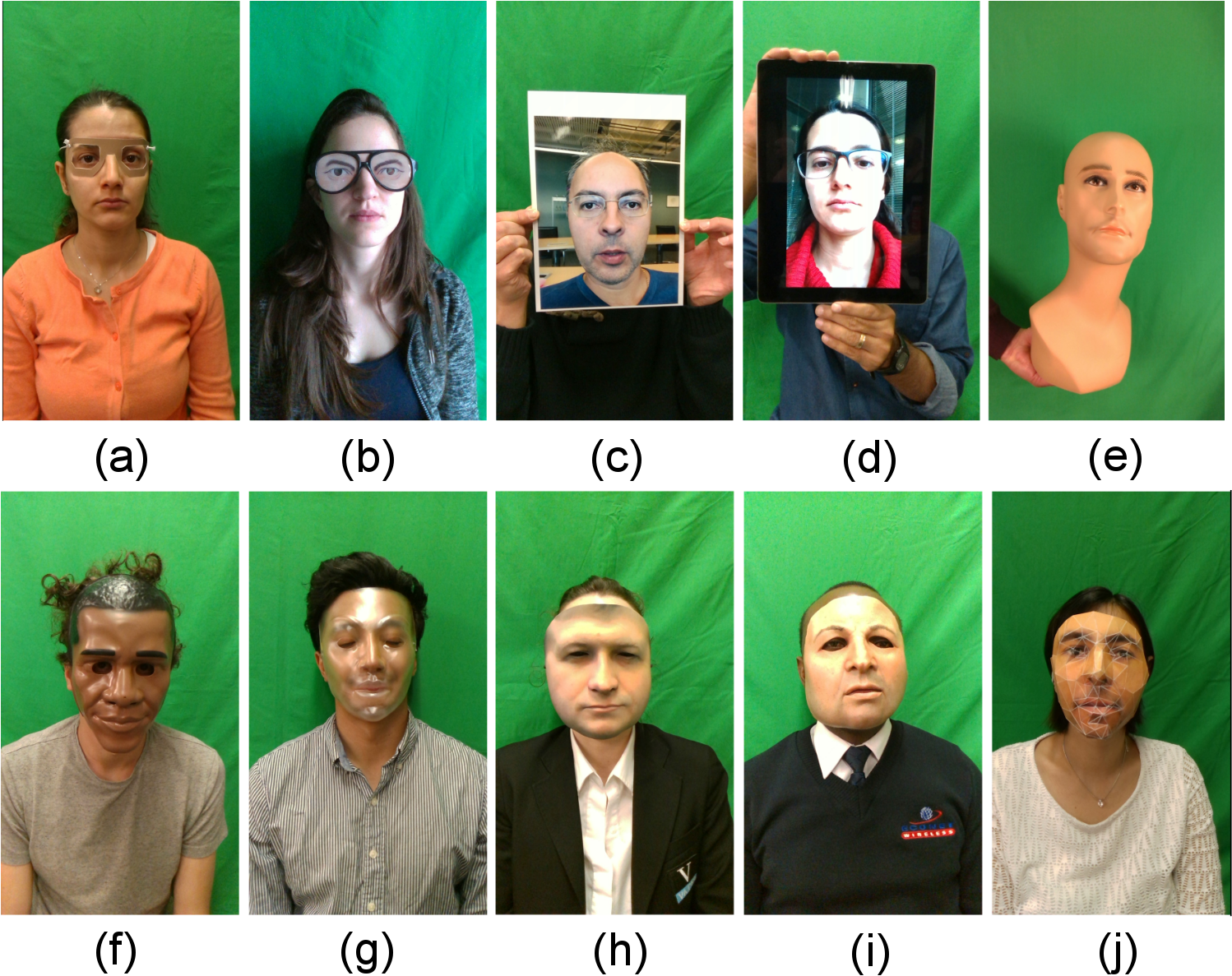}

\caption{Attack categories in \textit{WMCA}, (image taken from \cite{george_mccnn_tifs2019}) dataset (Color channel), (a): glasses (paper glasses), (b): glasses (funny eyes), (c): print (2D), (d): replay (2D), (e): fake head, (f): rigid mask (Obama mask), (g): rigid mask (transparent plastic mask), (h): rigid mask (custom made), (i): flexible mask (custom made), and (j): paper mask.}
\label{fig:pa_wmca}
\end{figure}

The key requirement of the proposed framework is the use of `multi-channel' information. 
We have conducted experiments in the publicly available  \textit{Wide Multi-Channel presentation Attack} (\textit{WMCA}) \cite{george_mccnn_tifs2019} database, which contains a wide variety of 2D and 3D presentation attacks. To the best of our knowledge, there are no other publically available PAD datasets that contain synchronized multi-channel data in a usable format for our task. One other notable multi-channel PAD dataset is CASIA-SURF \cite{zhang2020casia}. However, they only provide cropped and preprocessed images in the public distribution which cannot be used in an object detector framework. The \textit{WMCA} dataset contains a total of \textit{1679} video samples from \textit{72} individuals. Mainly, the database contains four different channels of information recorded simultaneously, namely, color, depth, infrared, and thermal channels, collected using two consumer-grade devices,  Intel\textsuperscript{\textregistered} RealSense\texttrademark SR300 (for color, depth and infrared), and Seek Thermal CompactPRO (for the thermal channel). The database contains different types of 2D and 3D attacks, such as print, replay, funny eyeglasses, fake head, rigid mask, flexible silicone mask, and paper masks. The statistics of attack types is shown in Table \ref{tab:BATL-data}. More details of the database can be found in \cite{george_mccnn_tifs2019}. Even though four different channels are available in the database, we use only three channels coming from the consumer-grade Intel RealSense camera (color, depth, and infrared). This makes the practical deployment more feasible as compared to the method in \cite{george_mccnn_tifs2019}. The samples of RGB images for the attack categories are shown in Fig. \ref{fig:pa_wmca}

\begin{table}[ht]
\centering
\caption{Number of samples in each presentation attack category in \textit{WMCA} database.}
\label{tab:BATL-data}
\begin{tabular}{lcr}
\toprule
Attack Type                        & Category & \#Videos                \\ \midrule
\textit{Bonafide} & -    & 347                            \\
Print                              & 2D   & 200                            \\
Replay                             & 2D   & 348                            \\
Fake head                          & 3D   & 122                            \\
Rigid mask                         & 3D   & 137                            \\
Flexible mask                      & 3D   & 379                            \\
Paper mask                         & 3D   & 71                             \\
Glasses                            & 3D (Partial)   & 75                    \\ \hline
\textbf{TOTAL}    &      & \textbf{1679} \\ \bottomrule
\end{tabular}
\end{table}

\subsection{Protocols}

We use the same \textit{grandtest} protocol defined in \cite{george_mccnn_tifs2019} for our experiments. However, it is to be noted that we only use three channels out of the four channels available from the database. All the channels we used are coming from the same consumer grade device making it suitable for deployment.

\subsection{Metrics}
For the evaluation of the algorithms, we have used the ISO/IEC 30107-3 metrics \cite{ISO}, Attack Presentation Classification Error Rate (APCER), and Bonafide Presentation Classification Error Rate (BPCER) along with the Average Classification Error Rate (ACER) in the $eval$ set. We compute the threshold in the $dev$ set for a BPCER value of 0.2\%. EPC curves and APCER and BPCER are also calculated. Additionally, the APCER of individual attack types and the ACER-AP which is defined as the average of BPCER and the maximum APCER of the attack types are also reported.

\subsection{Baseline methods}

We have implemented feature-based and CNN based baselines to compare with the proposed method. We used the same channels, i.e., gray-scale, depth, and infrared channels in most of the baselines for a fair comparison (except for the use of color channel in some baselines). We compared the proposed approach with competing multi-channel baselines using the same channels. The description of the baseline methods is given below.

\begin{itemize}
\item \textit{Haralick Fusion}: Here we use an extension of RDWT-Haralick-SVM method in \cite{agarwal2017face}. First, we perform a preprocessing stage, which consists of face detection. After that, the face region in all channels are normalized to the size of $128\times128$ pixels and rotated to make the eye-line horizontal. For each channel, Haralick \cite{haralick1979statistical} features computed from $4 \times 4$ grid are concatenated to form the feature vector. This feature vector was used with logistic regression (LR) for individual channels for the PAD task. A score fusion of all the channels was performed to get the final PAD score.

\item \textit{IQM-LBP fusion}: In this baseline, we followed a similar preprocessing as the previous baseline. After the preprocessing stage, features are extracted from different channels, for example, Image Quality Measures (IQM)~\cite{galbally2014image} were extracted for the RGB channel, and variants of Local Binary Patterns (LBP) features for non-RGB channels. For each channel, a logistic regression model is trained, and score level fusion is performed to obtain the PAD scores.

\item \textit{FASNet (Color)}: This is an RGB only CNN \cite{lucena2017transfer} baseline for the PAD task. Similar to the other baselines, this network also utilizes aligned images from the preprocessing stage.

\item \textit{MCCNN-GDI}: Here we implemented \cite{george_mccnn_tifs2019}, which achieved state of the art performance in the \textit{grandtest} protocol in \textit{WMCA} dataset, when used with gray-scale, depth, infrared and thermal channels. Here, to make it comparable with that of the proposed system, we retrained the system with the channels used in this work, i.e., gray-scale, depth, and infrared channels.

\end{itemize}

Additionally, for the feature-based methods, the best performing method from each channel is also added in the evaluations.

\subsection{Experiments}

\subsubsection{Baseline results}

The results of different baselines systems with color, depth, and infrared channels, as well as the fusion methods, are shown in Table \ref{tab:baseline_results}. For the individual channels, only the result from the best one is reported. Of all the channels present, the infrared channel performs the best with an ACER of 11.0\%. Though fusion improves the overall ROC, the performance in low BPCER regions becomes slightly worse. Out of the baselines, the MCCNN-GDI method is clearly superior, achieving an ACER of 3\%.

\begin{table*}[t]
\centering
\caption{Performance of the baseline systems and the components in \textbf{grandtest} protocol of \textit{WMCA} dataset. The values reported are obtained with a threshold computed for BPCER 0.2\% in $dev$ set.}
\label{tab:baseline_results}
\begin{tabular}{@{}lcc|ccc@{}}
\toprule
\multirow{2}{*}{Method} & \multicolumn{2}{c|}{dev (\%)}     & \multicolumn{3}{c}{test (\%)}                                  \\ \cmidrule(l){2-6}
                        & \multicolumn{1}{c|}{APCER} & ACER & \multicolumn{1}{c|}{APCER} & \multicolumn{1}{c|}{BPCER} & ACER  \\ \midrule

Color (Haralick-LR)         &41.1  &20.6  &42.0     &1.0  &21.5 \\
Depth (Haralick-LR)         &42.2  &21.2  &50.6  &0.2  &25.4 \\
Infrared (Haralick-LR)       &24.9  &12.5  &21.9  &0.0  &11.0 \\ \hline
Haralick Fusion  &29.0 &14.6 &27.0 &0.3 &13.6 \\
IQM-LBP fusion  &41.9 &21.0 & 47.7 & 0.3 &24.0 \\\hline
FASNet (Color) &29.3  &14.7  &24.9  &3.8  &14.3 \\
MCCNN-GDI &2.3 & 1.2 & 6.1 & 0.0 & 3.0 \\ \hline
\textbf{Proposed} &1.2 & 0.7 & 1.9 & 1.0 & \textbf{1.5} \\ \bottomrule
\end{tabular}
\end{table*}

\subsubsection{Results with the proposed framework}

\begin{figure}[h]
     \centering
         \includegraphics[width=1\linewidth]{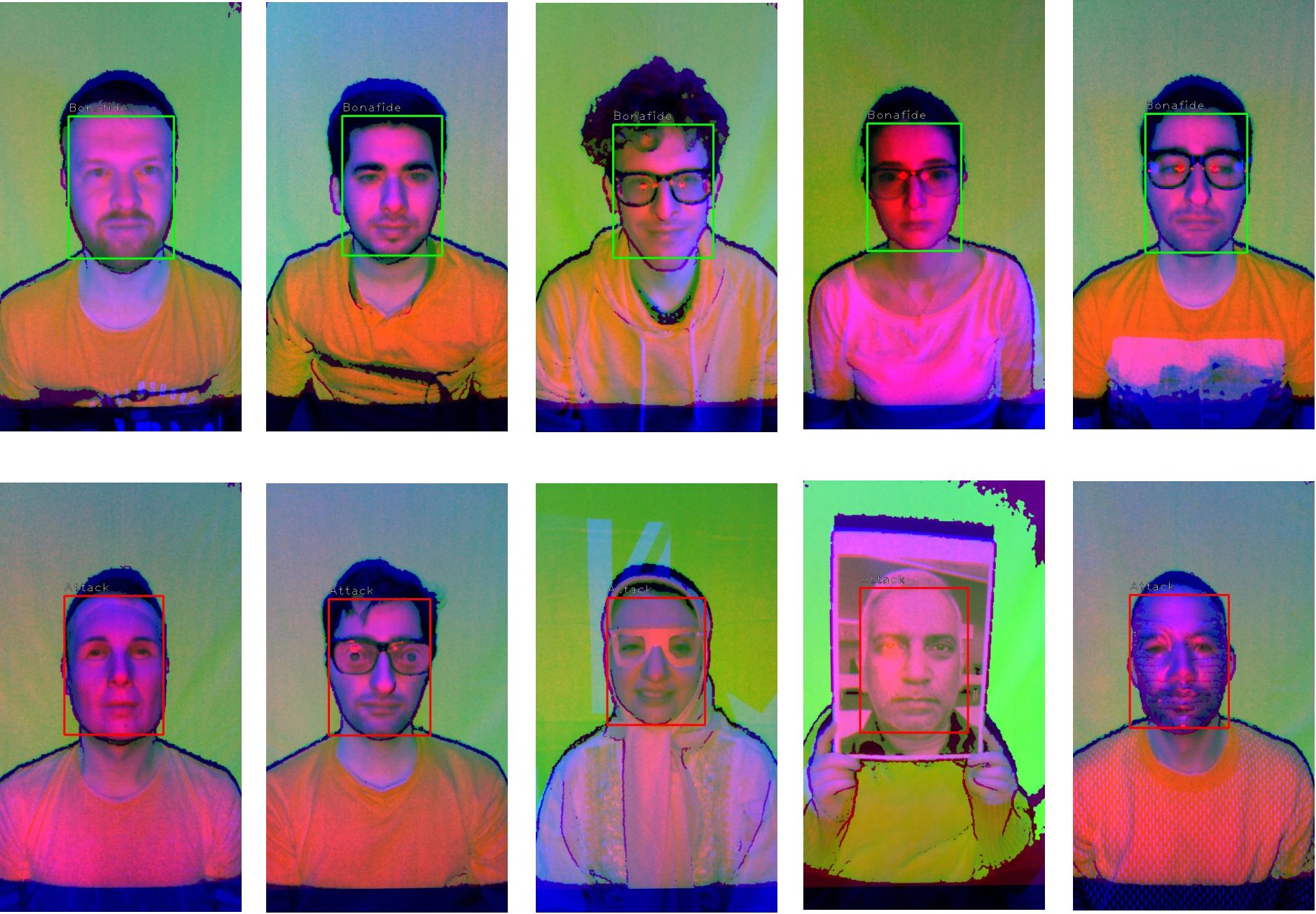}

\caption{ Image showing the successful detections, green boxes indicate the \textit{bonafide} detections and red boxes denote attack detections.
First row shows the successful \textit{bonafide} detections and second row shows the successful presentation attack detections.}
\label{fig:successful}
\end{figure}

From Table \ref{tab:baseline_results}, it can be seen that the proposed face detection based PAD framework achieves an ACER of 1.5\% outperforming the state of the art baselines. Some of the successful detections are shown in Fig. \ref{fig:successful}.

\begin{figure}[h]
     \centering
         \includegraphics[width=0.8\linewidth]{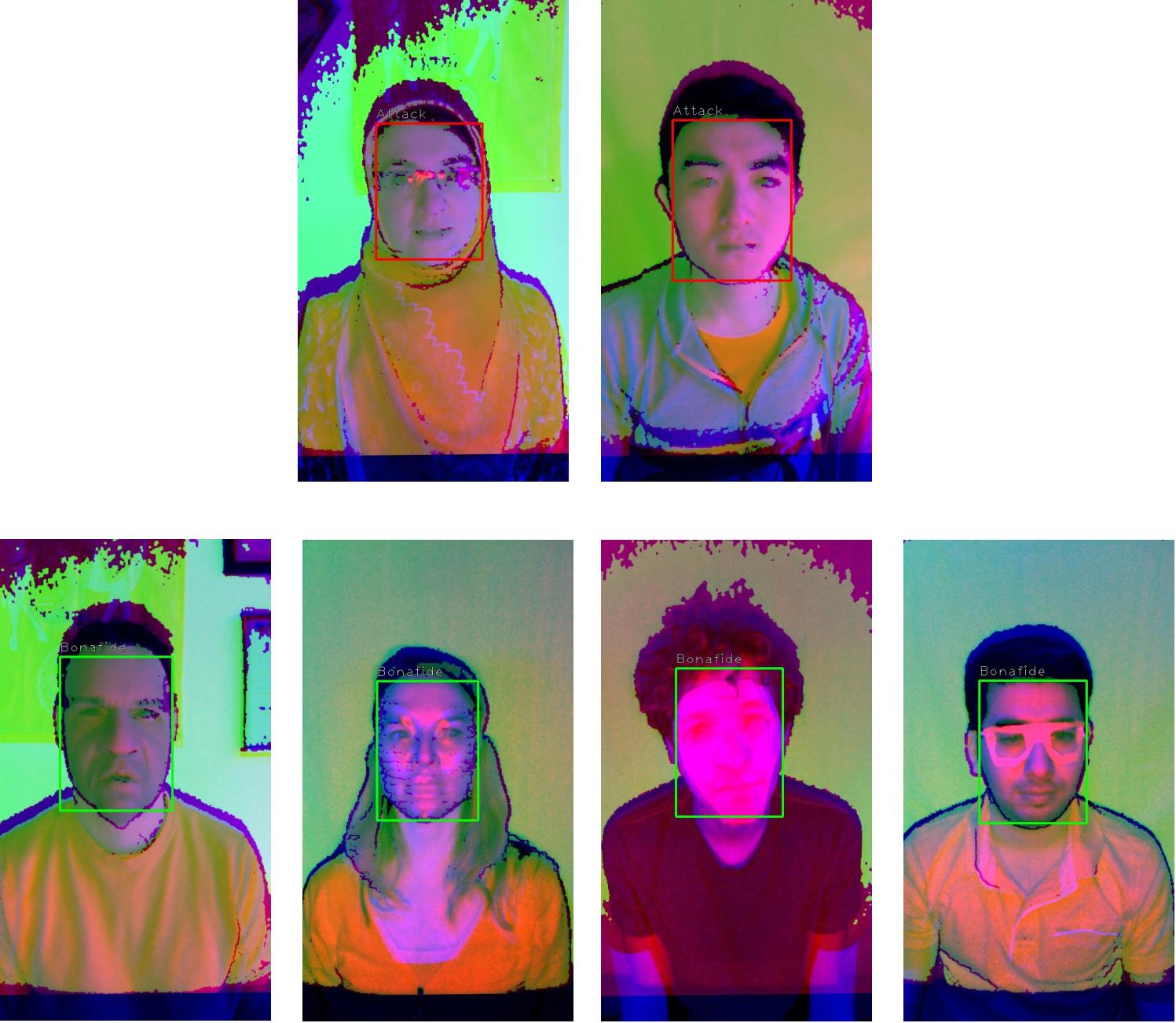}

\caption{ Image showing the failure case in detections, green boxes indicate the \textit{bonafide} detections and red boxes denote attack detections. The first row shows the failure cases where \textit{bonafide} are classified as attacks and the second row shows the failure cases where attacks are misclassified as \textit{bonafide}.}
\label{fig:failure}
\end{figure}

To identify the limitations of the framework, some of the false detections are also shown in Fig. \ref{fig:failure}. The instances of \textit{bonafide} getting classified as the attack could be due to the missing values in the depth channel. Depth data from the device contains holes where depth is not estimated correctly. This can probably be avoided with a preprocessing stage with hole filling and smoothing of the depth channel.

The same could explain the misclassification of the transparent mask as \textit{bonafide}. Some of the attacks with paper glasses are misclassified since it looks identical to \textit{bonafide} with corrective glasses.

\subsubsection{Detailed analysis of the scores}

In this section, we perform a detailed analysis of the scores to get insights into the misdetections and failure cases.
For this analysis, we compute the decision threshold in the development set with BPCER 0.2\% criterion, and we apply this threshold on the evaluation set. After that, we aggregate the APCER for individual attack categories. The results for different attack categories are tabulated in Table \ref{tab:detailed_pa}.

\begin{table}[h]
\centering
\caption{Detailed analysis of the performance for different PA's, at threshold computed for BPCER 0.2\% in $dev$ set.}
\label{tab:detailed_pa}
\begin{tabular}{@{}lccc@{}}
\toprule
Attack Type       & Category & Dev Set & Eval Set \\ \midrule
Prints        &2D& 0.0\%   & 0.0\%    \\
Replay        &2D& 0.0\%   & 0.0\%    \\
Flexible mask &3D & 2.0\%   & 0.8\%    \\
Rigid mask    &3D& 3.7\%   & 3.2\%    \\
Fake head     &3D& 0.0\%   & 0.4\%    \\
Glasses      &3D (Partial)& 2.5\%   & 11.9\%   \\ \hline
APCER-AP         & & 3.7\%   & 11.9\%   \\
BPCER             & & 0.2\%   & 1.0\%    \\\hline
\textbf{ACER-AP}  & &1.9\%    &6.5\% \\\bottomrule
\end{tabular}
\end{table}

From the Table \ref{tab:detailed_pa}, it can be seen that the proposed method achieves perfect performance in 2D attack classification. This is expected since the information from the depth channel alone might suffice for this task. The performance in 3D masks are also satisfactory, considering the very low BPCER values. The glasses attacks are the most challenging ones, and they contribute to most of the misclassifications. The glasses category consists of `Paper glass` (Fig. \ref{fig:pa_wmca} (a)) and `Funny eyes` (Fig. \ref{fig:pa_wmca} (b)) sub categories. Moreover, these are the only partial attacks present in the database. Especially in the channels considered, the funny eyes and the paper glass appear very similar to \textit{bonafide} samples with medical glass. This results in misclassification of these attacks.

The ROC plots provide information about the performance in the evaluation set only. However, in realistic conditions, the decision threshold has to be set a priori. Expected performance curves (EPC) \cite{bengio2005expected} provide an unbiased estimate of the performance of  (two-class) classifiers. As opposed to the ROC curve, where the evaluation is done only on the test set, EPC combines both development and test group to obtain the expected performance. The EPC curve shows the performance changes as a trade-off between False Matching Rate (FMR) and False Non Matching Rate (FNMR). This trade-off is controlled by the parameter \mbox{$0 \leq \alpha \leq1$} as follows:
\begin{equation}
\label{eqn:weighted_error}
WER = \alpha\times(FMR) + (1 - \alpha)\times(FNMR)
\end{equation}
where $WER$ (weighted error rate) is computed as a weighted combination of FMR and FNMR.

For each value of $\alpha$, the score-threshold which minimize the WER is selected based on the development group. The threshold chosen is used to compute the Half Total Error Rate (HTER) in the evaluation set.

\begin{figure}[ht]
\centering
\includegraphics[width=0.8\linewidth,page=5]{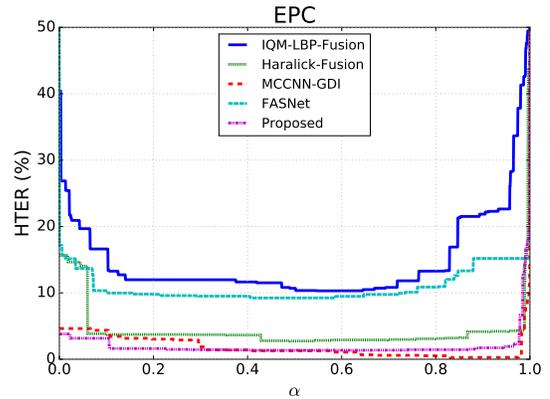}
\caption{EPC plots for the reproducible baselines and proposed method in the \textit{eval} set of \textit{grandtest} protocol in \textit{WMCA} dataset }
\label{fig:epc_WMCA}
\end{figure}

Fig. \ref{fig:epc_WMCA} shows the EPC plots of the proposed approach and the baseline methods. From the EPC curves, it can be seen that, in the lower BPCER regions (\mbox{$\alpha \leq0.5$}), the proposed approach outperforms the state of the art methods, indicating the robustness of the approach.

\subsection{Discussions}

From the experimental results, it can be seen that the proposed framework for PAD with multi-channel data works better than binary classifier based approaches in the low BPCER range. This framework is ideal for typical usage settings since face detection is a part of the face recognition pipeline. Instead of using conventional face detection methods which detect and pass both \textit{bonafide} and attacks to the subsequent stages, the proposed method uses a multi-channel face detector, specifically trained to localize and distinguish \textit{bonafide} and \textit{attack} ('\textit{non-face}') classes simultaneously. Also, it is worth noting that the hardware used in the proposed pipeline is a single off-the-shelf Intel RealSense SR300 camera. To sum it up, the proposed approach provides face PAD as a face detector itself. Currently, the multi channel presentation attack databases available are collected in controlled conditions. Availability of data collected `in the wild' conditions could make it possible to train the face detector using just the \textit{bonafide} samples. The harder negatives from the background might improve the performance of the detector greatly.

\section{Conclusions}

The proposed method shows a simple yet efficient method for PAD using multi-channel information. The method does not add additional complexity to a face recognition pipeline
since it can be a drop-in replacement for the face detector in the preprocessing stage. Even though the method is multichannel, the channels used are readily available from cheap consumer devices which makes it feasible for deployment. The Intel RealSense family of devices, Kinect and the upcoming OpenCV AI Kit (OAK-D) \cite{oak_d} are suitable candidates for such system as this model can be replaced with RGB-D channels as well. 

Addition of a PAD module is essential to secure face recognition systems against spoofing attempts. Typical PAD modules adds another layer of computational complexity to the face recognition pipeline. In this work, we propose a simple yet effective solution to add PAD capabilities to a face recognition system without adding any additional computational complexity. This is done by swapping the face detector in typical FR pipelines with the proposed multichannel face detector. The (multi-channel) face detector is used \textit{as} a presentation attack detector, thereby reducing the redundancy in the face recognition pipeline. To our best knowledge, this is the first method to use multi-channel face detection \textit{as} a PAD system. Specifically, a multi-channel face detector is trained to localize and classify \textit{bonafide} faces and presentation attacks. The proposed method utilizes color, depth, and infrared channels available from a commercially available sensors for PAD. The proposed method was compared with feature-based methods and state of the art CNN based methods for comparison in the \textit{WMCA} dataset and was found to outperform the state of the art method while obviating the requirement of additional face detection stage.

\section*{Acknowledgment}

Part of this research is based upon work supported by the Office of the
Director of National Intelligence (ODNI), Intelligence Advanced Research
Projects Activity (IARPA), via IARPA R\&D Contract No. 2017-17020200005.
The views and conclusions contained herein are those of the authors and
should not be interpreted as necessarily representing the official
policies or endorsements, either expressed or implied, of the ODNI,
IARPA, or the U.S. Government. The U.S. Government is authorized to
reproduce and distribute reprints for Governmental purposes
notwithstanding any copyright annotation thereon.

{\small
\bibliographystyle{IEEEtran}
\bibliography{egbib}
}

\end{document}